\documentclass[10pt,twocolumn,letterpaper]{article}

\usepackage[pagenumbers]{cvpr} 

\usepackage{tabularx}
\usepackage{float}
\usepackage{graphicx}
\usepackage{amsmath,mathrsfs}
\usepackage{amssymb}
\usepackage{booktabs}
\usepackage{multirow}
\usepackage{bbding}
\usepackage[accsupp]{axessibility}
\usepackage[linesnumbered,ruled,vlined,lined,boxed,commentsnumbered]{algorithm2e}

\definecolor{gray}{rgb}{0.92, 0.92, 0.92}
\definecolor{yellow}{rgb}{1, 1, 0.7}
\definecolor{orange}{rgb}{1, 0.85, 0.7}
\definecolor{red}{rgb}{1, 0.7, 0.7}

\definecolor{cvprblue}{rgb}{0.21,0.49,0.74}
\usepackage[pagebackref,breaklinks,colorlinks,allcolors=cvprblue]{hyperref}

\def\paperID{8344} 
\def\confName{CVPR}
\def\confYear{2025}

\title{CoSurfGS:Collaborative 3D Surface Gaussian Splatting with Distributed Learning for Large Scene Reconstruction}

\author{Yuanyuan Gao\footnotemark[1]~~$^{1}$, Yalun Dai\footnotemark[1]~~$^{2}$, Hao Li\footnotemark[1]~~$^{1}$, Weicai Ye\footnotemark[2]~~$^{3,4}$ \\
{Junyi Chen$^4$}, {Danpeng Chen$^{3}$}, Dingwen Zhang\footnotemark[2]~~$^{1}$, Tong He$^4$, Guofeng Zhang$^3$, Junwei Han$^1$\\
$^{1}$Brain and Artificial Intelligence Lab, Northwestern Polytechnical University \\
~~$^2$Nanyang Technological University ~~$^3$Zhejiang University
~~$^4$Shanghai AI Lab\\
\small \texttt{\{gyy7645,maikeyeweicai,zhangdingwen2006yyy\}@gmail.com}\\
\small \texttt{\{dialogue\_dylan, lifugan\_10027\}@outlook.com}\\
}

\begin{document}
\twocolumn[{
 \renewcommand\twocolumn[1][]{#1}
\maketitle
 \thispagestyle{empty}
 \pagestyle{empty}
 \begin{center}
     \captionsetup{type=figure}
    \vspace{-2em} \includegraphics[width=1\linewidth]{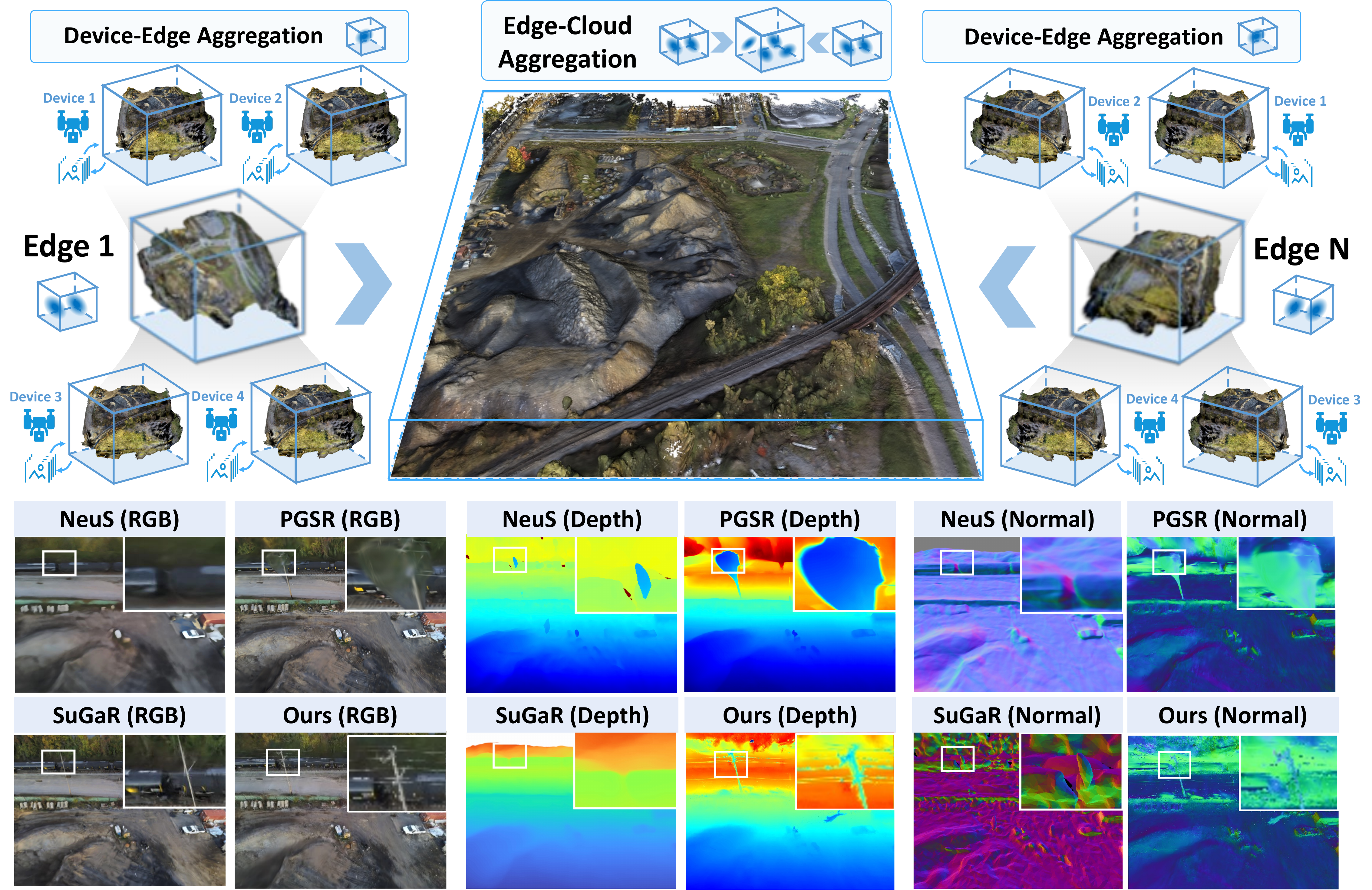}
    \vspace{-2em}
     \captionof{figure}{Our proposed \textbf{CoSurfGS} serves as a "device-edge-cloud" distributed learning framework that enables multi-agent parallel training. Under this framework, we can achieve superior large-scene reconstruction performance w.r.t the novel view synthesis, depth rendering, and surface normal prediction results (see the bottom part). Meanwhile, this framework can also accelerate the whole modeling process while preserving the privacy of local regions. }
     \label{fig:overview}
 \end{center}
 }]
\renewcommand{\thefootnote}{\fnsymbol{footnote}}
\footnotetext[1]{Equal Contribution.  \footnotemark[2]Corresponding author.}

\begin{abstract}
3D Gaussian Splatting (3DGS) has demonstrated impressive performance in scene reconstruction. However, most existing GS-based surface reconstruction methods focus on 3D objects or limited scenes. Directly applying these methods to large-scale scene reconstruction will pose challenges such as high memory costs, excessive time consumption, and lack of geometric detail, which makes it difficult to implement in practical applications. To address these issues, we propose a multi-agent collaborative fast 3DGS surface reconstruction framework based on distributed learning for large-scale surface reconstruction. Specifically, we develop local model compression (LMC) and model aggregation schemes (MAS) to achieve high-quality surface representation of large scenes while reducing GPU memory consumption. Extensive experiments on Urban3d, MegaNeRF, and BlendedMVS demonstrate that our proposed method can achieve fast and scalable high-fidelity surface reconstruction and photorealistic rendering. Our project page is available at \url{https://gyy456.github.io/CoSurfGS}.
\end{abstract}    
\section{Introduction}
\label{sec:intro}
The emergence of 3D Gaussian Splatting (3DGS)~\cite{kerbl3Dgaussians} has significantly revolutionized novel view synthesis (NVS), achieving both high fidelity and remarkable improvements in training and rendering speeds. It has rapidly been adopted as a general 3D representation across various tasks, including 3D scene perception~\cite{yu2024mip,yan2024multiscale3dgaussiansplatting}, dynamic scene reconstruction~\cite{yan2024streetgaussiansmodelingdynamic,wu20244dgaussiansplattingrealtime}, simultaneous localization and mapping (SLAM)~\cite{yan2024gsslamdensevisualslam,keetha2024splatamsplattrack,hu2024cgslamefficientdensergbd}, 3D generation and editing~\cite{zhou2024dreamscene360unconstrainedtextto3dscene,chen2024gaussianeditor,yi2024gaussiandreamerfastgenerationtext,liu2024humangaussiantextdriven3dhuman,wang2024gaussianeditorediting3dgaussians}.


However, 3DGS struggles to accurately represent 3D surfaces, primarily due to the inherent multi-view inconsistency of 3D Gaussians~\cite{huang20242d}, thus hindering its usage in areas like autonomous vehicles and urban planning.
To address this challenge, some recent works have extended 3DGS for surface reconstruction by flattening 3D Gassians into oriented elliptical disks and adding some multiview geometric constraints~\cite{huang20242d,chen2024pgsr}. 
Unfortunately, they would inevitably result in {low geometric accuracy}, {high memory costs}, and {excessive time consumption} when dealing with real-world large-scale scenes. To address these issues, we build a brand new framework, called CoSurfGS, for large-scale surface reconstruction with the following three-fold considerations. 


\textbf{High-quality surface.}
Ensuring high-quality surface reconstruction of large scenes is challenging as a single global model can hardly capture every geometric detail of the scene structure. To this end, in our framework, we convert the surface geometric optimization problem from a direct global-scene optimization to a progressive process from the local region to the global scene. For optimizing the surface geometry of each local region, we introduce single-view geometric constraints and multi-view geometric constraints to obtain local 3DGS models. Then, to gradually aggregate the surface structure from the local region to the global scene, we design a Model Aggregation Scheme (MAS), which adopts a self-knowledge distillation mechanism to maintain the key structure of each local region and align the surface geometry of the adjacent co-visible regions. These two issues are the key to obtaining the high-quality surface of large scenes.

\textbf{Low memory cost.}
For reconstructing the surface of large scenes, another critical issue is the memory cost of the whole process. To address this issue, we adopt Local Model compression (LMC) 
to each local model before aggregating it to the global scene. This is based on the finding that in most cases, the local models would have overlapping regions and contain lots of redundant Gaussian points themselves. To reduce such redundancy, we define a priority score to screen the Gaussian points lacking multi-view consistency and having low opacity.

\textbf{High-speed training.}
In addition to the memory cost, the time consumption of the large-scene reconstruction is always unbearable as well. So, how to speed up the whole training process is of great interest. In our framework, this problem is solved by the established  
distributed framework, which enables both the parallel 3DGS initialization and the parallel 3DGS training on each device. Such a framework greatly reduces the latency caused by data transmission. In addition, the aforementioned designs in MAS and LMC can further accelerate the speed of the final global model. 

We conduct both quantitative and qualitative evaluations on two datasets. 
Extensive experiment results highlight the superior rendering quality and impressive surface reconstruction performance of our approach. 
Fig. \ref{fig:overview} presents the framework, novel view synthesis, depth reconstruction, and surface reconstruction results. Our contributions are summarized as follows:

\begin{itemize}
\item We propose a collaborative large-scale surface reconstruction method based on distributed learning, achieving a substantial reduction in training time.
\item We propose the Local Model Compression (LMC) and Model Aggregation Scheme (MAS) for high-quality global scene surface representation with a lower GPU memory consumption.
\item 
Comprehensive experiments demonstrate that our method achieves state-of-the-art performance in surface reconstruction, surpassing all existing methods. It also delivers competitive results in novel view synthesis. Additionally, our CoSurfGS significantly reduces both training time and memory cost compared to all existing methods.
\end{itemize}

\section{Related Works}
\label{sec:related}

\subsection{Surface Reconstruction} 

Surface reconstruction is a fundamental task in computer vision and graphics, essential for producing high-fidelity 3D models from sparse or noisy input data. Traditional methods follow a multi-view stereo (MVS) pipeline, leveraging representations such as point clouds~\cite{lhuillier2005quasi}, volumes~\cite{kutulakos2000theory}, or depth maps~\cite{schoenberger2016mvs, campbell2008using}, but often suffer from artifacts due to erroneous matching and noise~\cite{barnes2009patchmatch}. 
Recent advances incorporate deep learning techniques~\cite{wang2021patchmatchnet, sarlin2019coarse} or employ neural representations like implicit fields~\cite{Park:2019} and occupancy grids~\cite{Niemeyer:2020} to enhance reconstruction quality, though computational complexity remains a challenge. Neural Radiance Fields (NeRF)~\cite{mildenhall2021nerf} have shown impressive results in rendering but struggle with capturing precise surface geometry, prompting further refinement through techniques such as Neus~\cite{wang2021neus} and GOF~\cite{yu2024gaussian}. 
To address these limitations, recent Gaussian Splatting approaches~\cite{chen2024pgsr, guedon2023sugar, huang20242d} decompose 3D Gaussian shapes into simpler forms.
However, existing methods do not consider computation efficiency and resource consumption, leading to excessively long training times for large-scale scene applications and even out-of-memory issues when computational resources are limited.

\subsection{Large Scale Reconstruction}
Traditional approaches~\cite{agarwal2011building, fruh2004automated, schonberger2016structure} follow a structure-from-motion (SfM) pipeline that estimates camera poses and generates sparse point clouds. However, such methods often contain artifacts or holes in areas with limited texture or speculate reflections as they are challenging to triangulate across images. 
Recently, NeRF~\cite{mildenhall2021nerf} and 3DGS~\cite{kerbl20233d} variants have become a worldwide 3D representation system thanks to their photo-realistic characteristics and the ability of novel-view synthesis, which inspires many works~\cite{zhenxing2022switch, zhang2023efficient, turki2022mega, tancik2022block, lin2024vastgaussian, xiangli2022bungeenerf, xu2023grid, suzuki2024fed3dgs} to extend it into large-scale scene reconstructions. 
The above methods can be categorized into centralized~\cite{zhenxing2022switch}, distributed.
%
Centralized methods (Grid-NeRF~\cite{xu2023grid}, GP-NeRF~\cite{zhang2023efficient}, etc.) adopt the integration of NeRF-based and grid-based methods to model city-scale reconstruction. 
Distributed methods (VastGaussian~\cite{lin2024vastgaussian}, Mega-NeRF~\cite{turki2022mega}, etc.) apply scene decomposition for multiple NeRF / Gaussian models optimization, 
%
However, with the growing scene size, all these methods limit their scalability due to the central server's limited data storage and unacceptable computation costs. Meanwhile, they all only focus on over-fitting the photo-realistic rendering but ignore the geometry performance. 
%
%
\section{Preliminaries}
\noindent \textbf{3D Gaussian Splatting (3DGS)}.
3DGS~\cite{kerbl3Dgaussians} represents a 3D scene as a collection of 3D Gaussian primitives $\mathbb{G} = \{\mathbf{G}_k\}$, where each Gaussian primitive is defined as: 
\begin{equation}
     \mathbf{G}_k({\mathbf{x}}|{\mu}_k, \mathbf{\Sigma}_k) = e^{-\frac{1}{2}(\boldsymbol{\mathbf{x}} - {\mu}_k)^\top\mathbf{\Sigma}_k^{-1}({\mathbf{x}}-\mathbf{\mu}_k)},
\end{equation}
where $\mathbf{\mu_k} \in \mathbb R^3$ is the center of the 3D Gaussian primitive, and ${\mathbf{\Sigma}_k} \in \mathbb R^{3\times3}$ is the 3D covariance matrix, which can be decomposed into the rotation matrix $\mathbf{R}_k\in \mathbb R^{3\times3}$ and the scaling matrix $\mathbf{S}_k\in \mathbb R^{3\times3}$ with $\mathbf{\Sigma}_k = \mathbf{R}_k\mathbf{S}_k\mathbf{S}_k^\top \mathbf{R}_k^\top$. The rendering process of the Gaussian is controlled by an opacity value \(o_k\) and the color value \(\textbf{c}_k\), the color is represented as a series of sphere harmonics coefficients in the practice of 3DGS, it facilitates the real-time alpha blending of numerous Gaussians to render novel-view images.

\noindent \textbf{PGSR}.
3DGS solely relies on image reconstruction loss, which lacks geometry accuracy, PGSR~\cite{chen2024pgsr} introduces geometric constraints based on the single-view consistency \(\mathcal{L}_{svg}\) and the multi-view consistency \(\mathcal{L}_{mvg}\). The former enforces that the normal vector \({\mathbf{N}_s}(\mathbf{p})\) calculated from the surrounding pixels is as same as possible to the normal vector \(\mathbf{N}(\mathbf{p})\) rendered at the pixel \(\mathbf{p}\):  
\begin{equation}
\label{eq:svg}
    \mathcal{L}_{svg} = \frac{1}{|\mathbb{W}|}\sum_{\mathbf{p}\in{\mathbb{W}}}{|\mathbf{N}_s(\mathbf{p})\mathbf{N}(\mathbf{p}) |}{||\mathbf{N}_s(\mathbf{p})-\mathbf{N}(\mathbf{p})||}_1,
\end{equation}
where \(|\mathbf{N}_s(\mathbf{p})\mathbf{N}(\mathbf{p})|\) considers the flatness around pixel \(\mathbf{p}\) to avoid the influence of edges, \(\mathbb{W}\) is the set of image pixels.

Then, it uses the homography matrix \({\mathbf{H}_{rn}}\) to keep the geometric multi-view consistency \(\mathcal{L}_{mvgeo}\)  and the photometric multi-view consistency \(\mathcal{L}_{mvrgb}\):
\begin{equation}
\left\{
\begin{aligned}
    \mathcal{L}_{mvgeo} &= \frac{1}{|\mathbb{V}|}\sum_{\mathbf{p}_r\in{\mathbb{V}}}||{\mathbf{p}_r-\mathbf{H}_{nr}\mathbf{H}_{rn}\mathbf{p}_r)}||,\\
    \mathcal{L}_{mvrgb} &= \frac{1}{|\mathbb{V}|}\sum_{\mathbf{p}_r\in{\mathbb{V}}}(1 -\text{NCC}(\mathbf{I}_r(\mathbf{p}_r),\mathbf{I}_n(\mathbf{H}_{rn}\mathbf{p}_r))),
\end{aligned}
\right.
\end{equation}


where \(\mathbf{p}_r\) denotes the pixel's 2D position in the reference frame, \(\mathbf{p}_n\) is the pixel projected by \(\mathbf{p}_r\) in the neighboring frame, \(\mathbf{I}_r(\mathbf{p}_r)\) and \(\mathbf{I}_n(\mathbf{H}_{rn}\mathbf{p}_r))\) denotes a certain size pixel patch centered at \(\mathbf{p}_r\) and \(\mathbf{p}_n\), \(\mathbf{NCC}(\cdot, \cdot)\) means the normalized cross correlation~\cite{yoo2009fast}, and when the reprojection error \(||(\mathbf{p}_r-\mathbf{H}_{nr}\mathbf{H}_{rn}\mathbf{p}_r)||\) exceeds a certain threshold \(\theta\), this pixel \(\mathbf{p}_r\) will be ignored, \(\mathbb{V}\) is the set of pixels whose error did not exceed the threshold. Finally, multi-view geometric constrains loss is ${\mathcal{L}_{mvg}} =  {\mathcal{L}_{mvgeo}}+{\mathcal{L}_{mvrgb}} $.

\begin{figure*}[t]
    \includegraphics[width=1\linewidth]{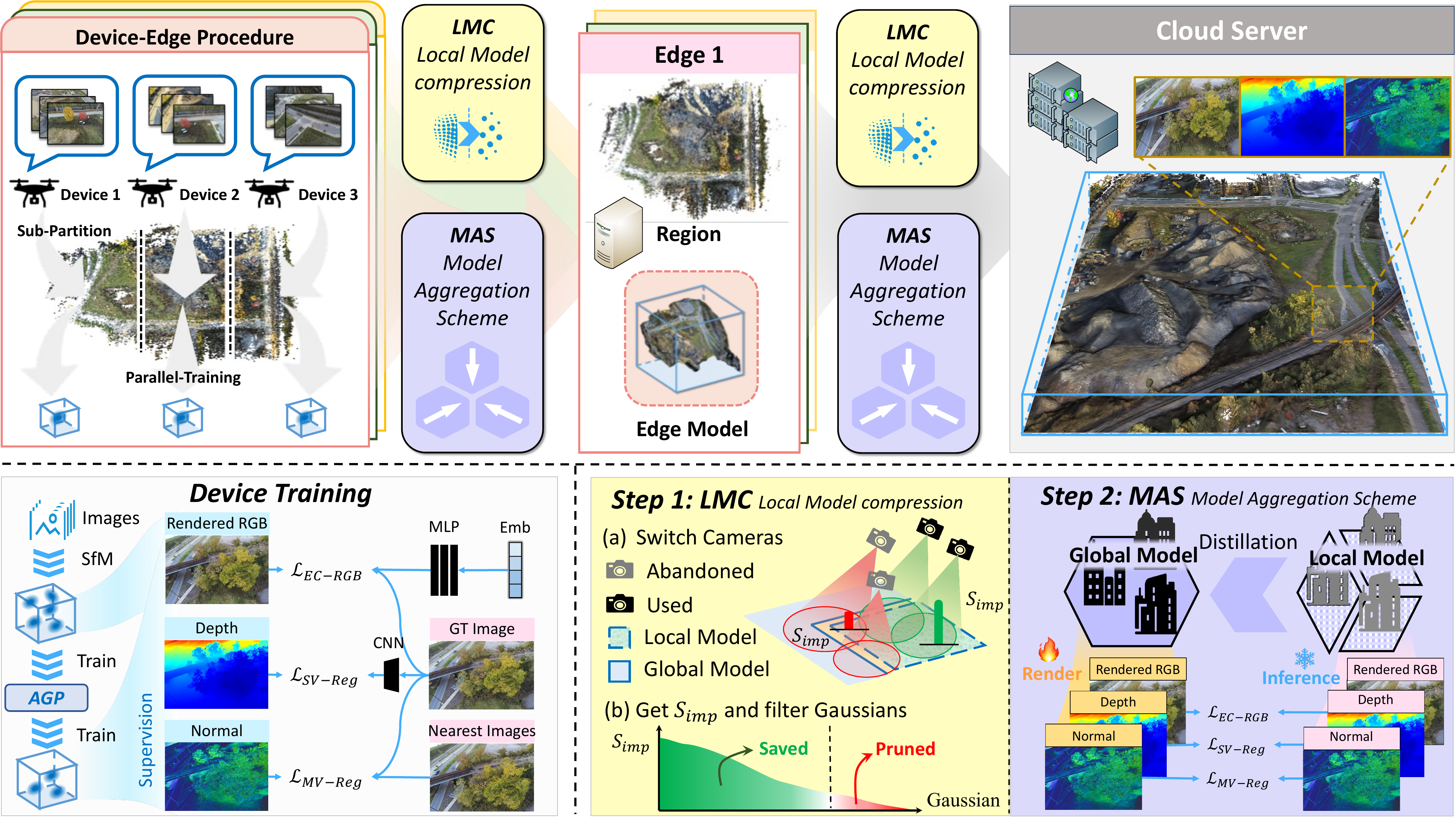}
    \caption{Our \textbf{CoSurfGS} follows "device-edge-cloud" three-layer distributed architecture. On the device side, each device is responsible for reconstructing an individual area by capturing images, performing SfM to initialize both extrinsic and intrinsic, and training the Gaussian models \(\mathbf{G}^L\). On the edge side, devices upload their Gaussian model to the edge followed by two-step aggregation techniques: 1) Local Model Compression (LMC) module prunes the redundant Gaussian points and abandons images with few contributions on reconstruction area; 2) Model Aggregation Scheme (MAS) module uses a self-distillation technique to aggregate the compressed model into a global model \(\mathbf{G}^G\). Moreover, the edge-cloud shares the same process as we've done on the edge side.}
    \label{fig:framework}
\end{figure*}

\section{Method}

\subsection{Overall Framework }
\label{sec: overall}
Our framework employs a device-edge-cloud architecture to enable distributed surface reconstruction.  In practice, each device (drone) captures a certain number of images and trains a Gaussian model \(\mathbb{G}^{d}_{i,j}\), \(i\) and \(j\) denote the \(i\)-th device in the \(j\)-th edge \( i\in [1, M], j\in[1, N]\). After that, a device-edge aggregation procedure is performed to aggregate the \(M\) nearby device models into the \(j\)-th edge Gaussian model by:
\begin{equation}
\label{eq:5}
    \mathbb{G}^{e}_{j} = f_{\text{MAS}}\left(f_{\text{LMC}}\left(\mathbb{G}^{d}_{1,j}\right), \cdots, f_{\text{LMC}}\left(\mathbb{G}^{d}_{M,j}\right)\right),
\end{equation}
where \(f_{\text{LMC}}\) operates the Local Model Compression process (see Sec.\ref{sec: local}) and \(f_{\text{MAS}}\) uploads the local models from devices to the edge server and aggregate them to obtain the \(j\)-th edge model \(\mathbb{G}^{e}_{j}\) (see Sec.\ref{sec: aggregation}). 

Following the same strategy, we apply \(f_{\text{MAS}}\) and \(f_{\text{LMC}}\) to the edge-cloud aggregation procedure for distributed large-scale scene reconstruction. Then we obtain:
\begin{equation}
\label{eq:6}
    \mathbb{G}^{c} = f_{\text{MAS}}\left(f_{\text{LMC}}\left(\mathbb{G}^e_1\right), \cdots, f_{\text{LMC}}\left(\mathbb{G}^e_{N}\right)\right).
\end{equation}
Notably, by only uploading Gaussian models instead of raw images, the framework can effectively guarantee each device's privacy. 
It can also be observed that both of the processes of Eq.\eqref{eq:5} and Eq.\eqref{eq:6} involve the identical transition from a relative local scale to a relative global scale. Thus, in subsequent sections, we will refer to the input of the device-edge/edge-cloud aggregation procedure as local model \(\mathbb{G}^l\), while the output as global model \(\mathbb{G}^g\).

\subsection{Device-side Training Procedure}
\label{sec: device}
Given \textit{i}-th device, inspired by PGSR~\cite{chen2024pgsr}, to transform 3D Gaussians into a 2D flat plane representation to accurately represent the geometry surface of the actual scene, we directly minimize the minimum scale factor \(\mathbf{S}_i=\text{diag}(\mathbf{s}_1, \mathbf{s}_2, \mathbf{s}_3)\)for each Gaussian:
\begin{equation}
\label{eq:scale}
    \mathcal{L}_s = {||min(\mathbf{s}_1,\mathbf{s}_2,\mathbf{s}_3)||}_1.
\end{equation}
In the early learning stage, i.e., the first \(\tau\) training iterations, we focused solely on image reconstruction loss following the origin 3DGS \cite{kerbl3Dgaussians}  \({\mathcal{L}_{3dgs}}(\mathbf{I}, \mathbf{I}_{gt}) = (1-\lambda)\|\mathbf{I}-\mathbf{I}_{gt}\|_1 + \lambda{\mathcal{L}_{SSIM}}(\mathbf{I}-\mathbf{I}_{gt})\) and scale loss \({\mathcal{L}_s}\). Then, we have the loss function for the first training stage:
\begin{equation}
\label{eq:2}
    {\mathcal{L}_{1}} = \mathcal{L}_{3dgs} + \lambda_1\mathcal{L}_{s}.
\end{equation}
After the first training stage, we additionally introduce the geometric constraints of PGSR\cite{chen2024pgsr} to the training process. The used loss function contains two parts:
the single view geometric loss \(\mathcal{L}_{svg}\) and the multi-view geometric loss \(\mathcal{L}_{mvg}\). Here we obtain the final loss \(\mathcal{L}\) as:
\begin{equation}
\label{eq:2}
    \mathcal{L} = \mathcal{L}_{1} + {\lambda}_{2}^{(t)}\mathcal{L}_{svg} + {\lambda}_{3}^{(t)}\mathcal{L}_{mvg}.
\end{equation}

Considering that abruptly adding geometric loss would make the model hard to converge and affect the rendering quality as well, we introduce a smooth weighting strategy to gradually add geometric weight along with the training iteration, where the  geometric weight \({\lambda}_{i}^{(t)}\) is defined as:
\begin{equation}
\label{eq:2}
    {\mathbf{\lambda}}_{i}^{(t)} = \beta_i\times\frac{t-\tau}{T},(t > \tau,i=2,3),
\end{equation}
\(t\) is the index of the training iteration, \(T\) denotes the max training iteration, \(\beta_i\) is a hyperparameter.   

During the training process mentioned above, the Gaussian model will get larger and larger through densification~\cite{kerbl3Dgaussians}. 
Although the densification process can improve the rendering performance remarkably, it would significantly increase the redundancy of the Gaussian points as well as the memory costs.
To this end, we deploy an Adaptive Gaussian Pruning (AGP) strategy to reduce the over-parameterized point number and preserve the original accuracy.
Specifically, for each Gaussian point \(\mathbf{G}_k= (\mathbf{x}_k, \mathbf{\Sigma}_k, \mathbf{S}_k, \alpha_k), \mathbf{G}_k\in \mathbb{G}^d\), we associate the priority score  \(S_{pro,k}\)  with the frequency at which the Gaussian point is projected onto the field of view of the image plane:
\begin{equation}
\label{eq:prune}
    \begin{aligned}
    S_{pro,k} & = \sum^{MHW}_{i=1} \mathbb{1}(\mathbf{G}_k, \mathbf{p}_i) \cdot \alpha_k \cdot \gamma(\Sigma_k), \\
\gamma(\Sigma) & =\left(\mathrm{V}_{\text {norm }}\right)^\beta, \\
\mathrm{V}_{\text {norm }} & =\min \left(\max \left(\frac{\mathrm{V}(\boldsymbol{\Sigma})}{\mathrm{V}_{\max 90}}, 0\right), 1\right),
\end{aligned}
\end{equation}
where \(M, H, W\) denotes the number, height, and width of the image. \(\mathbb{1}(\cdot, \cdot)\) is an indicator function determining whether a Gaussian point intersects with a given ray from a certain pixel. Here, \(\gamma(\Sigma)\) is used to provide an adaptive way to measure the dimension of its volume. It first normalizes the \(90\%\) largest of all sorted Gaussian and clips the range between 0 and 1. In this way, a higher priority score obtained by Eq.\ref{eq:prune} indicates Gaussian point can be projected to many image planes with a large size and high opacity, while a lower score indicates it's in the boundary of the scenes lacking multi-view consistency with a small size and low opacity. Consequently, we can easily prune the Gaussian points by introducing a hyperparameter \(\varphi\). 

\subsection{Local Model Compression}
\label{sec: local}
To ease the GPU consumption on edge / cloud, and let the global model optimize well in a limited training epoch, it is necessary to reduce the point redundancy of the local models, especially in regions near the other local models, before transiting them to the edge / cloud. 
Moreover, these redundant Gaussians trained by the local model lack accurate geometry representation due to the lack of multi-view consistency, which results in blurry and inconsistent geometry in certain areas.

To this end, we propose Local Model Compression \(f_{\textbf{LMC}}\), which compresses $\mathbb{G}^l$ to $\hat{\mathbb{G}}^l$. 
To remove redundant points accurately, before fusion, we establish another prune ratio \(\Psi\), determined by the proportion of cameras overlapping with the global model relative to those in the local model:
\begin{equation}
\label{eq:prune_percentage}
    {\Psi}\ = |\mathbb{C}^L\cap\mathbb{C}^g| / |\mathbb{C}^l|,
\end{equation}
where \(\mathbb{C}^l=\{\mathbf{K}^l_k,\mathbf{E}^l_k\},\mathbb{C}^g=\{\mathbf{K}^g_k,\mathbf{E}^g_k\}\) denotes a set of cameras from the local model and global model, \(\mathbf{K}^l_k,\mathbf{E}^l_k\) is the intrinsic and extrinsic matrix of camera. We utilize Eq.(\ref{eq:prune}) to compute the priority ranking of Gaussian points in the local model under camera viewpoints that do not overlap with those of the global model \(\hat{\mathbb{C}}^l\):
\begin{equation}
\label{eq:pl}
    \hat{\mathbb{C}}^l = \mathbb{C}^l-\mathbb{C}^l\cap\mathbb{C}^g,
\end{equation}
then remove Gaussian points with the lowest \({\Psi}\) of scores.

\subsection{Model Aggregation Scheme }
\label{sec: aggregation}
To aggregate the local models \(\hat{\mathbb{G}}^l_i\) into the global model \(\mathbb{G}^g\), one intuitive idea is to merge all Gaussian points from local models as follows directly: $    \mathbb{G}^g = \hat{\mathbb{G}}^l_1\cup\hat{\mathbb{G}}^l_2\cup \cdots \cup \hat{\mathbb{G}}^l_M$. 
However, such a strategy results in noticeable blurring in the boundary regions between local models.

Previous centralized methods~\cite{lin2024vastgaussian,chen2024dogaussian} tackle this issue with a two-step optimization: expanding each local model’s training area and trimming and merging the models at the boundaries. However, expanding the training area increases training time and local device computational resources. Additionally, collecting images from adjacent regions for expansion may infringe on the privacy of neighboring devices.

\begin{table*}[h]
\centering
\caption{\textbf{Quantitative results of surface reconstruction on BlendedMVS dataset}.
 We present Precision, Recall, and F-Score metrics for our mesh evaluation.
 $\uparrow$: higher is better, $\downarrow$: lower is better.
          The \colorbox{red}{red}, \colorbox{orange}{orange} and \colorbox{yellow}{yellow} colors respectively 
          denote the best, the second best, and the third best results.}
\resizebox{1.0\textwidth}{!}{
          \small
\begin{tabular}{c|ccc|ccc|ccc|ccc}
\toprule
\multirow{2}{*}{Method}                                 & \multicolumn{3}{c|}{Scene-01} & \multicolumn{3}{c|}{Scene-02} & \multicolumn{3}{c|}{Scene-03} & \multicolumn{3}{c}{Scene-04} \\
                                                        & precision$\uparrow$       & recall$\uparrow$      & f-score$\uparrow$      & precision$\uparrow$       & recall$\uparrow$      & f-score$\uparrow$      & precision$\uparrow$       & recall$\uparrow$      & f-score$\uparrow$      & precision$\uparrow$       & recall$\uparrow$      & f-score$\uparrow$      \\ \midrule
PGSR                                                    & \cellcolor{orange}0.2632          & \cellcolor{orange}0.3148      & \cellcolor{orange}0.2867       & \cellcolor{yellow}0.2675          & \cellcolor{orange}0.4486      & \cellcolor{yellow}0.3351       & \cellcolor{red}0.3462          & \cellcolor{orange}0.3419      & \cellcolor{orange}0.3440       & \cellcolor{orange}0.6690          & \cellcolor{orange}0.7287      & \cellcolor{orange}0.6976       \\
NeUS                                                    & \cellcolor{yellow}0.2219          & 0.2223      & \cellcolor{yellow}0.2271       & \cellcolor{orange}0.3324          & 0.3447      & \cellcolor{orange}0.3384       & \cellcolor{yellow}0.1118          & \cellcolor{yellow}0.1229      & \cellcolor{yellow}0.1171       & \cellcolor{yellow}0.1786          & \cellcolor{yellow}0.3819      & \cellcolor{yellow}0.2434       \\
BakedAngelo & 0.2190          & \cellcolor{yellow}0.1780      & 0.1964       & 0.2006          & \cellcolor{red}0.4735      & 0.2818       & 0.0949          & 0.1055      & 0.0999       & 0.0837          & 0.2634      & 0.1270       \\
CoSurfGS (Ours)                                                   & \cellcolor{red}0.2768          & \cellcolor{red}0.3152      & \cellcolor{red}0.2947       & \cellcolor{red}0.3600          & \cellcolor{yellow}0.4419      & \cellcolor{red}0.3967       & \cellcolor{orange}0.3459          & \cellcolor{red}0.3486      & \cellcolor{red}0.3472       & \cellcolor{red}0.6900          & \cellcolor{red}0.7346      & \cellcolor{red}0.7116       \\ \bottomrule
\end{tabular}
}
\label{tab:geometery}
\end{table*}

\begin{table*}[t!]
 \centering
   \caption{\textbf{Quantitative results of novel view synthesis on Mill19~\cite{turki2022mega} dataset 
          and UrbanScene3D~\cite{lin2022capturing} dataset}. $\uparrow$: higher is better, $\downarrow$: lower is better.
          The \colorbox{red}{red}, \colorbox{orange}{orange} and \colorbox{yellow}{yellow} colors respectively 
          denote the best, the second best, and the third best results.
          \textbf{Bold} denotes the best result in the 'With Mesh' group.
 }
    \resizebox{\linewidth}{!}{
        \LARGE
        \begin{tabular}{l ccc ccc ccc ccc ccc}
            \toprule[1.1pt]
             &   \multicolumn{3}{c}{\emph{Building}}  
             &   \multicolumn{3}{c}{\emph{Rubble}} 
             &   \multicolumn{3}{c}{\emph{Campus}} 
             &  \multicolumn{3}{c}{\emph{Residence}} 
             &   \multicolumn{3}{c}{\emph{Sci-Art}} \\
             \cmidrule(r){2-4} \cmidrule(r){5-7} \cmidrule(r){8-10} \cmidrule(r){11-13} \cmidrule(r){14-16} 
            &  SSIM$\uparrow$ & PSNR$\uparrow$ & LPIPS$\downarrow$   
            &  SSIM$\uparrow$ & PSNR$\uparrow$ & LPIPS$\downarrow$ 
            &  SSIM$\uparrow$ & PSNR$\uparrow$ & LPIPS$\downarrow$   
            &  SSIM$\uparrow$ & PSNR$\uparrow$ & LPIPS$\downarrow$   
            &  SSIM$\uparrow$ & PSNR$\uparrow$ & LPIPS$\downarrow$  \\
            \midrule
            \textbf{No mesh}  \\
            \midrule

            Mega-NeRF    
             & 0.547 & 20.92 & 0.454 
             & 0.553 & 24.06 & 0.508 
             & 0.537 & 23.42 & 0.636 
             & 0.628 & 22.08 & 0.401 
             & 0.770 & \cellcolor{orange}25.60 & 0.312 \\
            
            Switch-NeRF  
             & 0.579 & 21.54 & 0.397 
             & 0.562 & 24.31 & 0.478
             & 0.541 & 23.62 & 0.616 
             & 0.654 & \cellcolor{red}22.57 & 0.352 
             & 0.795 & \cellcolor{red}26.51 & 0.271 \\

            VastGaussian 
            & 0.728 & 21.80  & \cellcolor{yellow}0.225 
            & 0.742 & 25.20  & \cellcolor{red}0.264 
            & \cellcolor{orange}0.695 & \cellcolor{orange}23.82 & \cellcolor{red}0.329 
            & 0.699 & 21.01 & \cellcolor{yellow}0.261 
            & 0.761 & 22.64 & \cellcolor{orange}0.261 \\

            3DGS
            & \cellcolor{yellow}0.738 & \cellcolor{orange}22.53 & \cellcolor{orange}0.214 
            & 0.725 & \cellcolor{orange}25.51 & 0.316 
            & \cellcolor{yellow}0.688 & \cellcolor{yellow}23.67 & \cellcolor{orange}0.347 
            & \cellcolor{yellow}0.745 & \cellcolor{orange}22.36 & \cellcolor{orange}0.247 
            & 0.791 & 24.13 & \cellcolor{yellow}0.262 \\ 
            
            Hierarchy-GS
             & 0.723 & 21.52  & 0.297 
             & 0.755 & 24.64 & 0.284 
             & -- & -- & -- 
             & -- & -- & --
             & -- & -- & -- \\

            DOGS
            & \cellcolor{red}0.759 & \cellcolor{red}22.73 & \cellcolor{red}0.204
            & \cellcolor{yellow}0.765 & \cellcolor{red}25.78 & \cellcolor{red}0.257 
            & 0.681 & \cellcolor{red}24.01 & 0.377
            & 0.740 & 21.94 &  \cellcolor{red}0.244
            & \cellcolor{red}0.804 & \cellcolor{yellow}24.42 & 0.219 \\

            \midrule

            \textbf{With mesh}  \\
            \midrule

            PGSR
            & 0.480 & 16.12 & 0.573 
            & 0.728 & 23.09 & 0.334 
            & 0.399 & 14.02 & 0.721 
            & \cellcolor{orange}0.746 & 20.57 & 0.289 
            & \cellcolor{yellow}0.799 & 19.72 & 0.275 \\
            PGSR+VastGS
            & 0.720 & 21.63 & 0.300
            & \cellcolor{orange}0.768 & 25.32 & \cellcolor{yellow}0.274
             & -- & -- & -- 
             & -- & -- & --
             & -- & -- & -- \\
            
            SuGaR   
            & 0.507 & 17.76 & 0.455 
            & 0.577 & 20.69 & 0.453 
            & --    & --    & --
            & 0.603 & 18.74 & 0.406 
            & 0.698 & 18.60 & 0.349 \\

            NeuS     
             & 0.463 & 18.01 & 0.611 
             & 0.480 & 20.46 & 0.618 
             & 0.412 & 14.84 & 0.709 
             & 0.503 & 17.85 & 0.533 
             & 0.633 & 18.62 & 0.472 \\

            Neuralangelo & 0.582 & 17.89 & 0.322
                         & 0.625 & 20.18 & 0.314
                         & 0.607 & 19.48 & 0.373
                         & 0.644 & 18.03 & 0.263
                         & 0.769 & 19.10 & \cellcolor{red}0.231 \\

            \textbf{CoSurfGS (Ours)} & 
            \cellcolor{orange}\textbf{0.750} & \cellcolor{yellow}\textbf{22.40} & \textbf{0.262} & 
            \cellcolor{red}\textbf{0.774} & \cellcolor{yellow}\textbf{25.39} & \cellcolor{orange}\textbf{0.267} & 
            \cellcolor{red}\textbf{0.719} & \textbf{23.63} & \cellcolor{yellow}\textbf{0.360} & 
            \cellcolor{red}\textbf{0.776} & \cellcolor{yellow}\textbf{22.31} & \textbf{0.261} & 
            \cellcolor{orange}\textbf{0.802} & \textbf{23.29} & {0.277}\\
            \bottomrule[1.1pt]
        \end{tabular}
    }
    \label{tab: compare}
\centering
\end{table*}

To resolve boundary blurring without increasing computational costs, we adopt a self-distillation mechanism inspired by distributed learning \cite{lin2024vastgaussian, chen2024dogaussian} to optimize the global model. 
Firstly, each local uploads their compressed model \(\hat{\mathbb{G}}^l_i\) and corresponding cameras \(\mathbb{C}^l_i\) to initialize the global model $\mathbb{G}^g$.   
In order to largely maintain the rendering quality and the surface geometry accuracy, we use the local models as the teacher model, leveraging their RGB, normal, and depth maps to supervise and optimize our global model $\mathbb{G}^g$, since the Gaussian points are already sufficient, we do not perform densification, the optimization of global model $\mathbb{G}^g$ is as follows:
\begin{equation}
    \label{eq:simple-dist-obj}
    \begin{aligned}
        \underset{\mathbb{G}^g}{\arg\min} \; & \mathbb{E}_{(\mathbf{K}^{l}_k,\, \mathbf{E}^{l}_k)\sim\mathbb{C}^l_i} \Biggl[ \mathcal{L}_\text{g} \Bigl( R(\mathbf{K}^{l}_k, \mathbf{E}^{l}_k, \mathbb{G}^g), \\
        & \qquad\qquad\quad R(\mathbf{K}^{l}_k, \mathbf{E}^{l}_k, \hat{\mathbb{G}}^l_i) \Bigr) \Biggr],
    \end{aligned}
\end{equation}
where \(R(\cdot,\cdot)\) means the rendering process of RGB, depth, and normal image.
Same as the training on the device, we also use the scale loss \(\mathcal{L}_s\) and single-view loss \(\mathcal{L}_{svg}\) to form $\mathcal{L}_{g}$ to keep the Gaussian flat and the geometric consistency of a single image. Besides we further involve $\mathcal{L}_d$ and $\mathcal{L}_n$ to constrain the depth and normal of the large-scene surface, which is defined as:
\begin{equation}
\left\{
\begin{aligned}
    \mathcal{L}_d &= \|\mathbf{D}_g - \hat{\mathbf{D}}_l \|_1, \\
    \mathcal{L}_n &= \| \mathbf{N}_g - \hat{\mathbf{N}}_l \|_1,
\end{aligned}
\right.
\end{equation}
where \(\hat{\mathbf{D}}_l, \hat{\mathbf{N}}_l\) is the depth, normal map rendered by the local model, \(\mathbf{D}_g, \mathbf{N}_g\) denotes the depth, normal map rendered by the global model.
After optimization, we prune redundant Gaussians based on their opacity and size, leading to a refined global model.

\section{Experiments}

\begin{table*}[t]
\small
 \centering
 \caption{\textbf{Training Resources Consumption on Mill19 dataset and UrbanScene3D dataset}.
 We present the time (hh: mm), the number of final points ($10^6$), and the allocated memory (GB) during evaluation. For 3DGS-based methods, the overall training time includes the COLMAP process and training process.
 }
 \label{table:urban3d_quantitative_time}
 \begin{tabular}{l  l l l l  l l l l  l l l l  l l l }
 \toprule[1.1pt]
 
 \multirow{2}{*}{Models} &
 \multicolumn{2}{c}{Building} &
 \multicolumn{2}{c}{Rubble} & 
 \multicolumn{2}{c}{Campus} &
 \multicolumn{2}{c}{Residence} & 
 \multicolumn{2}{c}{Sci-Art} \\
 
 \cmidrule(r){2-3} \cmidrule(r){4-5} \cmidrule(r){6-7} \cmidrule(r){8-9} \cmidrule(r){10-11}

 & \multirow{1}{*}{Time $\downarrow$}
 & \multirow{1}{*}{Mem $\downarrow$} 
 
 & \multirow{1}{*}{Time $\downarrow$}
 & \multirow{1}{*}{Mem $\downarrow$} 
 
 & \multirow{1}{*}{Time $\downarrow$}
 & \multirow{1}{*}{Mem $\downarrow$} 

 & \multirow{1}{*}{Time $\downarrow$}
 & \multirow{1}{*}{Mem $\downarrow$} 

 & \multirow{1}{*}{Time $\downarrow$}
 & \multirow{1}{*}{Mem $\downarrow$} \\

 \midrule

 Mega-NeRF~\cite{turki2022mega}
 & 19:49  & 5.84 
 & 30:48  & 5.88 
 & 29:03  & 5.86 
 & 27:20  & 5.99 
 & 27:39  & 5.97 \\

 Switch-NeRF~\cite{zhenxing2022switch}
 & 24:46  & 5.84  
 & 38:30  & 5.87  
 & 36:19  & 5.85  
 & 35:11  & 5.94  
 & 34:34  & 5.92  \\

 $\text{3DGS}$~\cite{kerbl3Dgaussians}
 & 21:37  & 4.62  
 & 18:40  & 2.18 
 & 23:03  & 7.69  
 & 23:13  & 3.23  
 & 21:33  & 1.61  \\

 $\text{VastGS}^{\dagger}$~\cite{lin2024vastgaussian}
 & 04:14  & 3.07 
 & 04:00  & 2.74 
 & 07:24  & 9.61 
 & 04:59  & 3.67 
 & 04:51  & 3.54 \\

 DOGS~\cite{chen2024dogaussian}
 & 04:39  & 3.39 
 & 03:55  & 2.54 
 & 08:09  & 4.29 
 & 06:20  & 6.11 
 & 06:41  & 3.53  \\

PGSR+VastGS~\cite{chen2024pgsr,lin2024vastgaussian}
 & 05:30  & 3.14 
 & 04:30  & 3.15 
 & -  & - 
 & -  & - 
 & -  & -  \\
 
 \midrule
 CoSurfGS (Ours)
 & \textbf{03:49}  & \textbf{2.23}
 & \textbf{03:28}  & \textbf{2.63} 
 & \textbf{06:00}  & \textbf{2.22} 
 & \textbf{03:30}  & \textbf{2.35} 
 & \textbf{04:50}  & \textbf{1.06} \\ \bottomrule[1.1pt]
\end{tabular}
\end{table*}

\subsection{Settings}
\textbf{Baselines.} 
For large-scale scene NVS, we choose the mainstream NeRF-based and 3DGS-based methods, such as Mega-NeRF~\cite{turki2022mega}, Switch-NeRF~\cite{zhenxing2022switch}, VastGaussian~\cite{lin2024vastgaussian}, modified 3DGS~\cite{kerbl3Dgaussians}, and Hierarchy-GS~\cite{hyun2024adversarial} as our benchmark.
Moreover, for surface reconstruction, we use SOTA methods such as Neuralangelo~\cite{li2023neuralangelo}, NeuS~\cite{wang2021neus}, PGSR~\cite{chen2024pgsr}, and SuGaR~\cite{guedon2023sugar} as the comparative methods.

\noindent\textbf{Other details.} For details on the datasets and implementation, please refer to Supp. \ref{sec: exp dataset} and \ref{sec: exp implementation}.

\subsection{Main Results}

\begin{figure*}[htbp]
    \includegraphics[width=1\linewidth]{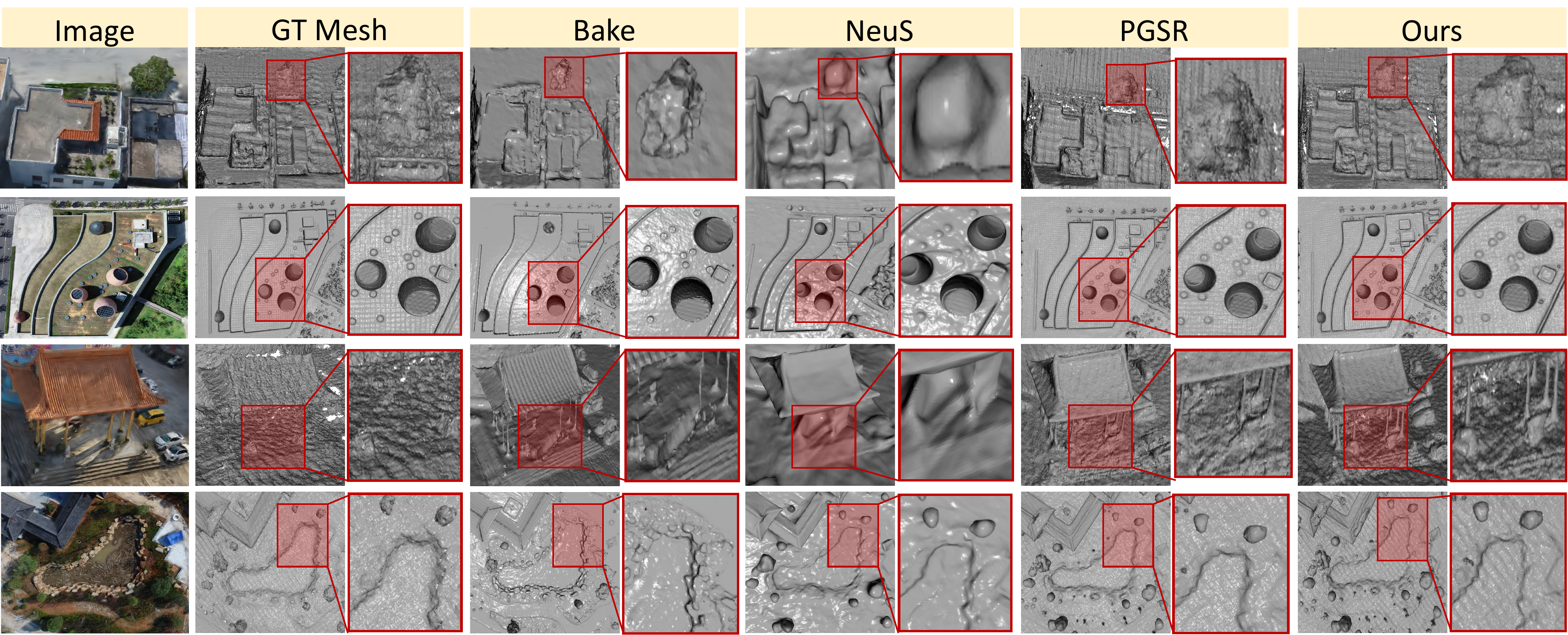}
    \caption{3D mesh comparison between our method and other surface reconstruction methods. The result of Scene-01, Scene-02, Scene-03, and Scene-04 are represented from top to bottom. The discriminate area are zoomed up by '\textcolor{red}{$\Box$}'.}
    \label{fig:mesh_visual}
\end{figure*}
\noindent \textbf{Surface Reconstruction} 
%
%
%
To evaluate the surface geometry accuracy of our method, we conduct experiments on large-scale scenes from the BlendedMVS dataset.
The quantitative results are shown in Tab.~\ref{tab:geometery}, where our method boosts the performance compared with existing reconstruction methods, by +0.05 in the F-score metric average.

\begin{figure*}[t]
    \includegraphics[width=1\linewidth]{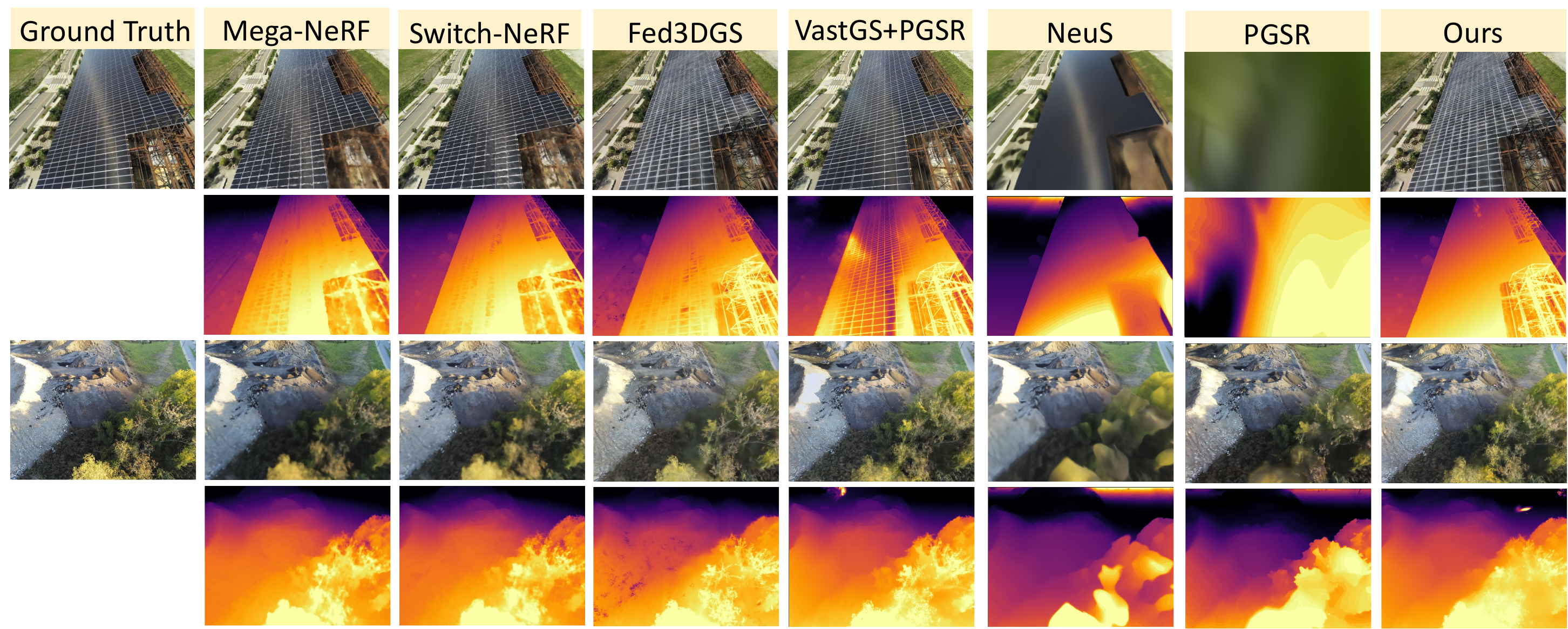}
    \caption{Qualitative results of our method and other methods in image and depth rendering, it shows the result of Rubble and Building, other large scenes visualization can be seen in the Supp.~\ref{sec: Additional depth}.}
    \label{fig:rgb_visual}
\end{figure*}

Moreover, we conduct some qualitative experiments to further evaluate the effectiveness of our method.
Following~\cite{zeng20173dmatch}, we visualize our 3D Mesh results and compare them with other methods, as shown in Fig.~\ref{fig:mesh_visual}. 
As the discriminate regions highlighted in the figure show, our method not only fully represents the entire scene, but also captures accurate details of the geometric representation.
%
Additionally, we provide the visualization results of rendered normal maps in Supp.~\ref{sec: Additional normal}. 
Take the "Rubble" scene as an example, we are the only method that can model the detail of the telegraph pole on both depth and RGB images.
\noindent \textbf{Novel View Synthesis}
In Tab.~\ref{tab: compare} and Fig.~\ref{fig:rgb_visual}, we quantitatively and qualitatively compare the rendering quality of the recent large-scale NVS (w/o mesh) and large-scale surface reconstruction (w/ mesh) methods.
Our method achieves state-of-the-art results in large-scale surface reconstruction and is comparable to large-scale NVS methods.

From the Tab.~\ref{tab: compare}, it's evident that we have significant improvement compared to the large-scale surface. We have an average advantage of +0.1, +3.0, and +0.4 in SSIM, PSNR, and LPIPS metrics.
The significant improvement we've achieved is due to our distributed approach, each device is responsible for a smaller region, allowing for better convergence. 
In contrast, other surface reconstruction methods are usually fully trained on the entire scene, leading to underrepresentation of the whole large-scale scene and resulting in suboptimal accuracy and rendering quality. Moreover, this often causes out-of-memory (OOM) errors that prevent the training process.
Fig.~\ref{fig:rgb_visual} demonstrates the lack of detail in the rendering results of these surface reconstruction methods, resulting in a blurry appearance.

\noindent \textbf{Training Resources Consumption}
Apart from impressive novel view synthesis and surface reconstruction performance, we additionally compare the training resources consumption with other large-scale reconstruction methods in Tab.~\ref{table:urban3d_quantitative_time}, the metrics include training time and memory costs. As can be seen, our method achieves the lowest time and memory cost, making large-scale scene surface reconstruction more practical and usable in real-world applications.

\subsection{Ablation Study}

\begin{figure}[t]
    \vspace{-1em}
    \includegraphics[width=1\linewidth]{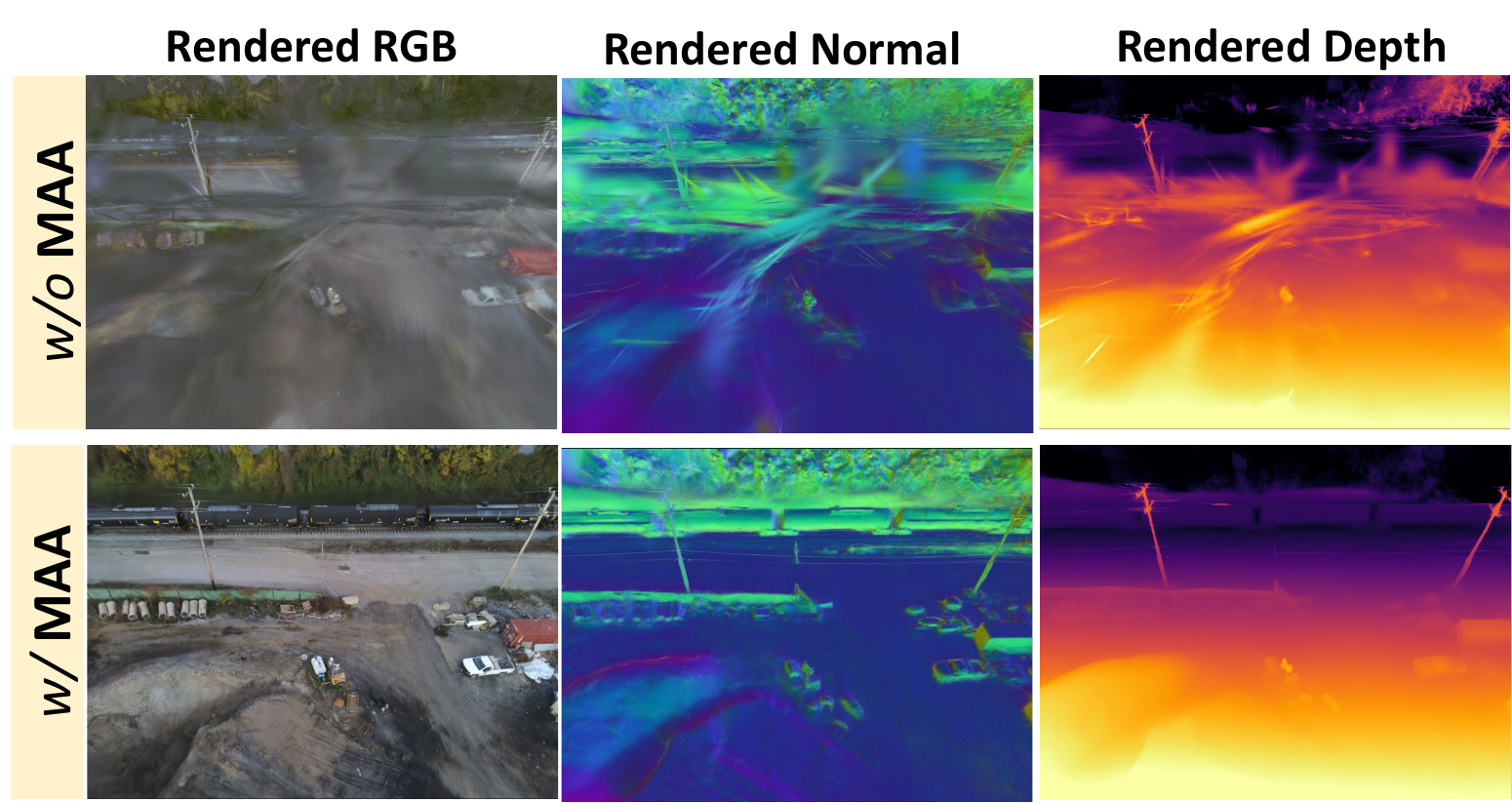}
    \caption{Qualitative results of Model Aggregation Scheme in Rubble dataset.}
    \label{fig:ablation_MAS}
\end{figure}

\begin{table}[t]
\centering
\footnotesize
\caption{Ablations of Model Aggregation Scheme.}
\begin{tabular}{l|cccc}
 \toprule
\multicolumn{1}{l|}{ \textbf{Model setting}} & PSNR$\uparrow$ & SSIM$\uparrow$ & LPIPS$\downarrow$ & Mem\\ \midrule
 w/ MAS&25.39&0.774 &0.267    &2.63GB  \\
 w/o MAS&18.07 &0.441&0.584 &2.74GB  \\  \bottomrule
\end{tabular}
\label{tab: MAS_ab}
\end{table}

\noindent \textbf{Model Aggregation Scheme(MAS).}
Tab.~\ref{tab: MAS_ab} demonstrates that including the MAS step in our method results in notable enhancements. The performance metrics improve by +7.0 in PSNR, +0.3 in SSIM, and +0.3 in LPIPS, compared to our method without MAS, and employing the MAS also results in a lighter model. Additionally, Fig.~\ref{fig:ablation_MAS} shows that \textit{w/o} MAS results in lots of floaters in the boundary regions, which indicates our MAS can mitigate blurring in the boundary regions between local models.

\begin{figure}[t]
\vspace{-1.em}
    \includegraphics[width=1\linewidth]{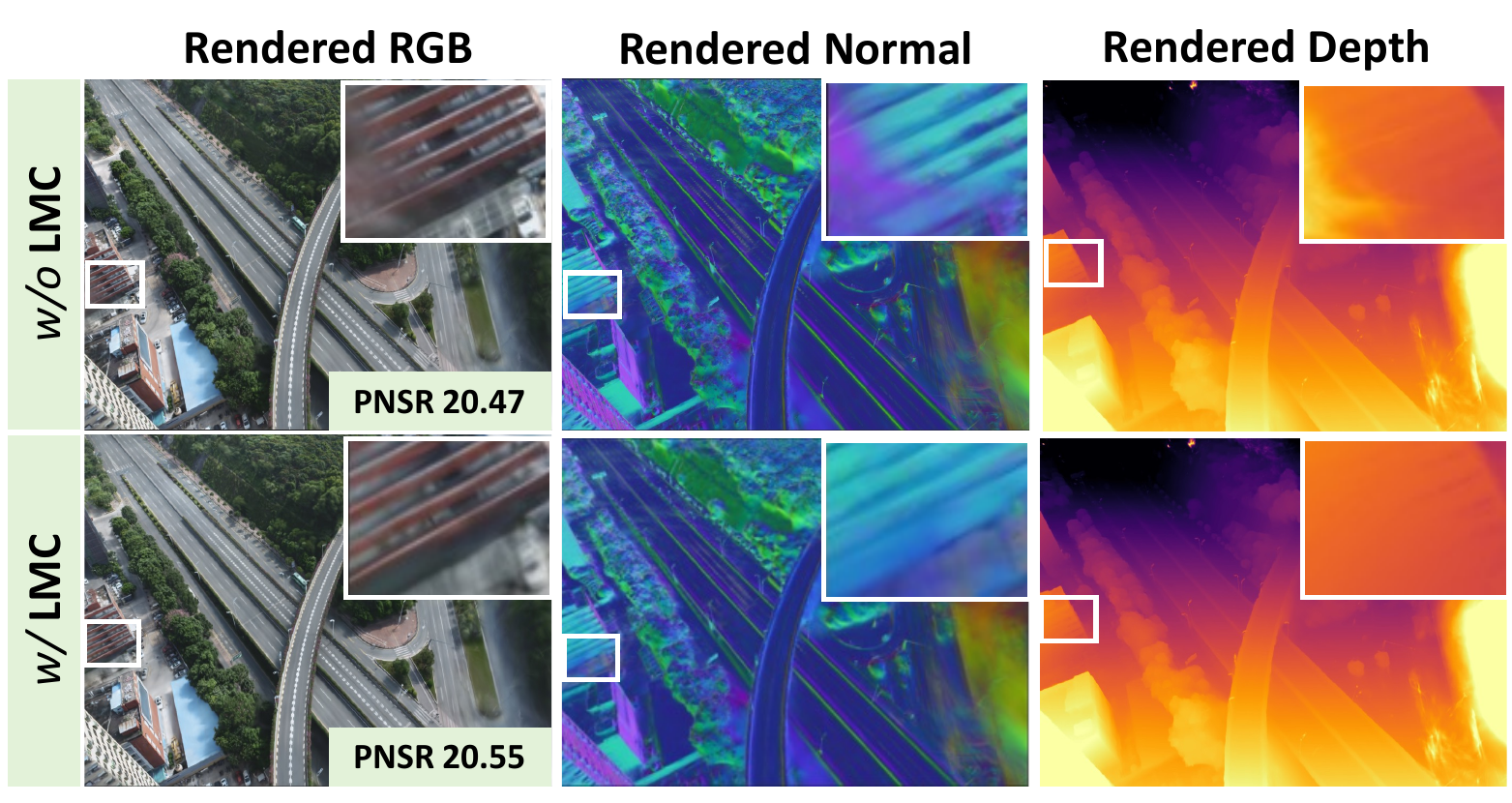}
    \caption{Qualitative results of Local Model Compression module in Residence dataset.}
    \label{fig:ablation_lmc}
    \vspace{-1em}
\end{figure}

\noindent \textbf{Local Model Compression (LMC). }
Tab.~\ref{tab: LMC_ab} shows that using LMC cuts the model memory by half and improves performance on the NVS, compared to directly using the local model without LMC. 
Fig.~\ref{fig:ablation_lmc} reveals that the inaccurate geometry representation of the aggregated model without LMC 
 would cause blurry areas in the rendered image and artifacts in the predicted normal and depth maps. 
This is because the redundant Gaussian points in the boundary areas always lack multi-view consistency, while our LMC eliminating these points leads to a more consistent normal and smoother depth representation on the plane surface. 

\begin{table}[t]
\centering
\caption{Ablations of Local Model Compression.}
\footnotesize
\begin{tabular}{c|cccc}
 \toprule
\multicolumn{1}{l|}{ \textbf{Model setting}} & PSNR$\uparrow$ & SSIM$\uparrow$ & LPIPS$\downarrow$ & Mem \\ \midrule
 w/ LMC&22.23&0.786 &0.316    & 475MB \\
 w/o LMC&22.13 &0.779&0.323 & 874MB \\  \bottomrule
\end{tabular}
\label{tab: LMC_ab}
\end{table}

\begin{table}[t]
\centering
\footnotesize
\caption{Ablations of prune percentage \(\varphi\).}
\begin{tabular}{c|cccc}
 \toprule
{\textbf{Prune percent}} & PSNR$\uparrow$ & SSIM$\uparrow$ & LPIPS$\downarrow$  & Mem\\ 
\midrule
40\%                                  & 25.19         & 0.765                                & 0.285                                & 2.5GB                                   \\
60\%                                  & 25.08                                & 0.752                                & 0.300                                & 1.68GB                                  \\
80\%                                  & 24.82                                & 0.724                                & 0.336                                & 1.16GB                                  \\ 
\bottomrule
\end{tabular}
\label{tab: prune}
\end{table}

\noindent \textbf{Compression percentage. }
Due to the limited storage and computational capabilities of edge devices, adopting our pruning strategy is crucial.
To further explore our proposed compression method, we tested three different compression percentages \(\varphi\) on the Rubble scene during device training.
%
While substantially pruning the Gaussians to get a lighter model, we strive to minimize the degradation of rendering metrics as much as possible. 
Tab.~\ref{tab: prune} indicates that a pruning rate of 80\% reduces the memory usage to less than half of that seen with a 40\% pruning rate, yet still maintains high rendering quality.

Other ablations can be seen in Supp.~\ref{sec: ab_geo} and ~\ref{sec: ab_patition} .
%

\section{Conclusions}
In this paper, we have proposed a  "device-edge-cloud” framework to enable distributed surface reconstruction.
For the device-edge and edge-device aggregation procedure, the proposed LMC module can eliminate the redundant Gaussians between local models, and the MAS module helps optimize the merged global models. Extensive experiments on the UrbanScene3D, MegaNeRF and BlendedMVS datasets demonstrate that 
our method achieves the highest surface reconstruction accuracy, shortest time, lowest memory cost, and has a comparable rendering quality compared to the current state-of-art methods.


\
{
    \small
    \bibliographystyle{ieeenat_fullname}
    \bibliography{main}
}
\
\clearpage
\setcounter{page}{1}
\maketitlesupplementary

\section{Additional Ablation Studies}
\subsection{Geometry Supervision in Device Training}
\label{sec: ab_geo}
\begin{figure}[b]
    \includegraphics[width=1\linewidth]{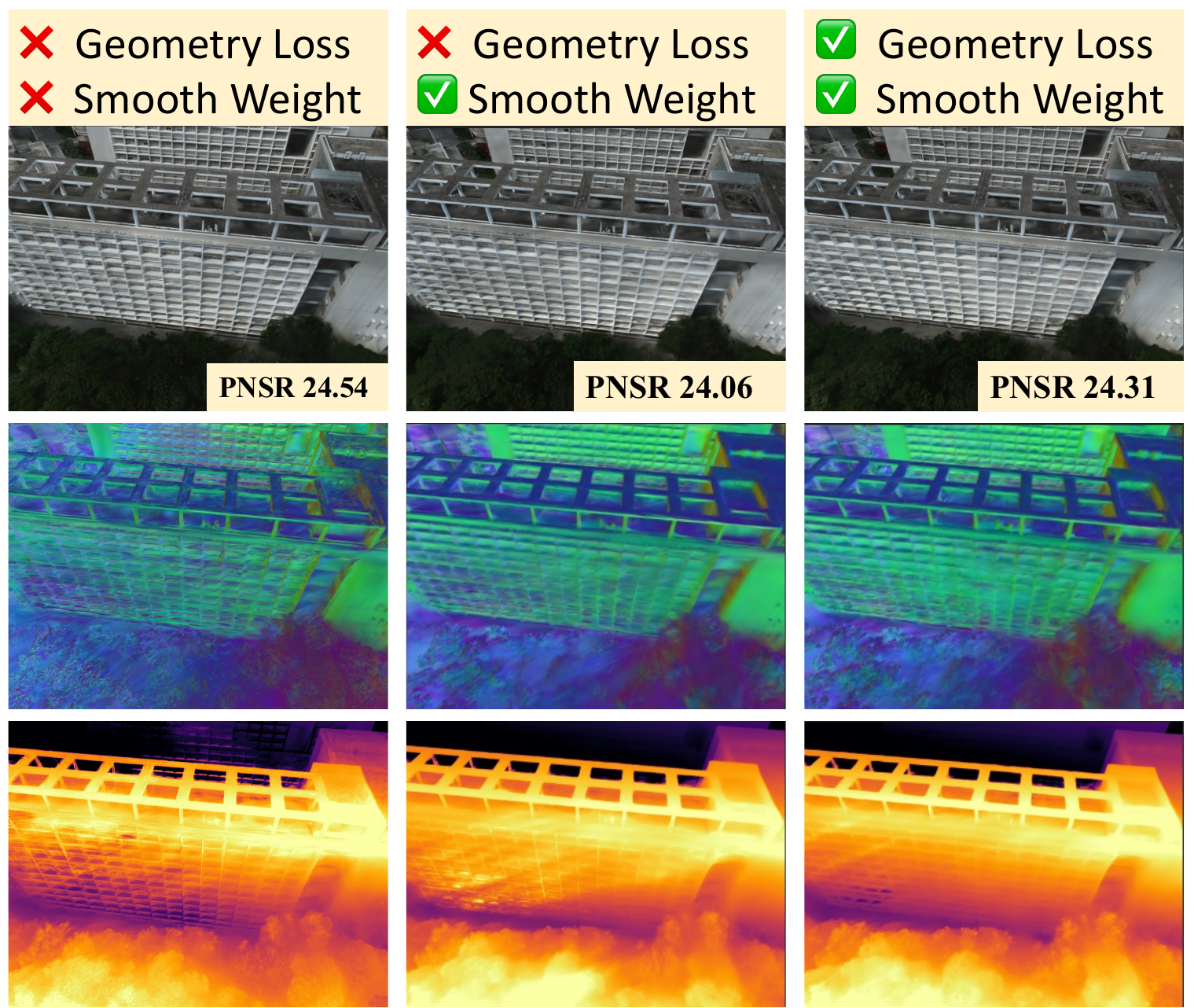}
    \caption{Qualitative results of our method and other methods, here we demonstrate the rendered RGB images, normal predictions, and depth estimations.}
    \label{fig:ablation_geometry}
\end{figure}

The performance of device training is critical for the final results since it is the teacher model of the distillation-based aggregation procedure. 
Therefore, we carefully design a geometry supervision strategy during the device training to improve both novel-view synthesis results and geometry accuracy.
Here we conduct ablations of both geometry loss and smooth weight, the quantitative results are shown in Tab. \ref{tab:ablation_geometry}, where the qualitative results are shown in Fig. \ref{fig:ablation_geometry}. 
It demonstrates that without geometry supervision, the device side performs poorly in geometry representation (\textit{e.g.} depth and normal prediction )

\subsection{Numbers of Edges and Devices}
\label{sec: ab_patition}
Unlike Vastgaussian~\cite{lin2024vastgaussian} Our method does not require strict limitations on the partition strategy. In our setting, we simply partition the large scene with uniform segmentation. In Tab. \ref{tab: partition}, we test different partition strategies on the Rubble datasets. (3*3)*4 means we have 4 edge models and one edge model contains 9 device models, So in total the large scene is divided into 36 blocks, it demonstrates that with more devices. The rendering quality will have a little decrease due to the more co-visible Gaussian points between devices that require redistribution, but it will gain the time advantage by the faster convergence in device training.

\section{Details}
\subsection{Datasets}
\label{sec: exp dataset}
For surface reconstruction evaluation, we choose four scenes in BlendedMVS~\cite{yao2020blendedmvs}, Scene-01, Scene-02, Scene-03, Scene-04 correspond to the scene 5bbb6eb2ea1cfa39f1af7e0c, 5b271079e0878c3816dacca4, 5b864d850d072a699b32f4ae, 5b60fa0c764f146feef84df0, each of them represents a relatively large scene and consists of more than 600 images. For photometric fidelity evaluation, we choose real-life aerial large-scale scenes, which encompass the \textit{Building} (1940 images) and \textit{Rubble} (1678 images) scenes extracted from Mill-19~\cite{turki2022mega}, and \textit{Campus} (5871 images), \textit{Residence} (2581 images), and \textit{Sci-Art} (3018 images) from the UrbanScene3D dataset~\cite{lin2022capturing}. Each scene contains thousands of high-resolution images. We downsample the images by 4 times for training and validation, following previous methods~\cite{turki2022mega}.

\subsection{Implementation Details}
\label{sec: exp implementation}
Our method is implemented using Pytorch, and all experiments are conducted on A100 40GB GPU.
For the partition from the cloud to the edge, we simply divide the cloud into 4 equal areas, except Building which has a relatively smaller area is divided into 2 equal areas.
From the edge to the devices, all scenes are divided into 4 equal areas. 
During the device training, the training iteration is set to 30,000, the prune iteration is set to 20,000 and the prune percentage \(\varphi\) is set to 0.2, image reconstruction loss weight \(\lambda\) = 0.2, and the scale loss weight \(\lambda_1\) = 25, after the first training stage(7000 iterations),
we set the max geometric weight \(\beta_2\) to 0.01, \(\beta_3\) consists the multi-view geometric constraints and multi-view photometric weights, each is 0.05 and 0.2, The patch size for Multi-view photometric loss is set from 11$\times$11 to 7$\times$7, and the threshold \(\theta\) is set to 1 to choose valid pixels. 
In MAS, the distillation epoch is set to be 5, depth loss weight and normal loss weight is set to 0.015.
What's more, the F-score  of Scene-01, Scene-02, Scene-03, Scene-04 is calculated under the error margin of 0.5, 0.1, 0.5, 0.2 meters.
\begin{table}[h]
\centering
\footnotesize
\caption{Ablations of Geometry Loss in Device-side Training.}
\begin{tabular}{cc|ccc}
 \toprule
 Geometry Loss & Smooth Weight & PSNR$\uparrow$ & SSIM$\uparrow$ & LPIPS$\downarrow$  \\ \midrule
 \XSolidBrush & \XSolidBrush &25.03 & 0.874 & 0.224   \\
 \checkmark & \XSolidBrush &24.36 & 0.864 & 0.240   \\
 \checkmark & \checkmark   &24.73 & 0.866 & 0.237   \\ \bottomrule
\end{tabular}
\label{tab:ablation_geometry}
\end{table}
\begin{table*}[t]
\centering
\caption{Ablations of different partition strategies.}
\begin{tabular}{
>{\columncolor[HTML]{FFFFFF}}l |
>{\columncolor[HTML]{FFFFFF}}l 
>{\columncolor[HTML]{FFFFFF}}l 
>{\columncolor[HTML]{FFFFFF}}l 
>{\columncolor[HTML]{FFFFFF}}l 
>{\columncolor[HTML]{FFFFFF}}l 
>{\columncolor[HTML]{FFFFFF}}l 
>{\columncolor[HTML]{FFFFFF}}l }
 \toprule
{\textbf{partition}} & PSNR$\uparrow$ & SSIM$\uparrow$ & LPIPS$\downarrow$  & {\textbf{Mem}} & {\textbf{cloud}} & {\textbf{device}} & {\textbf{edge}} \\ 
\midrule
{(1*1)*1}            & OOM         & {OOM}         &{OOM}         & {OOM}           & {OOM}          & {OOM}            & { OOM}    \\
{(1*1)*4}            & {25.37}         & {0.772}         & {0.280}         & {1.58GB}           & {70min}          & {0min}            & {130min}        \\
{(2*2)*4}            & {25.19}         & {0.765}         & {0.285}         & {2.5GB}            & {58min}          & {12min}           & {80min}         \\
{(3*3)*4}            & {24.97}         & {0.734}         & {0.325}         & {2.16GB}           & {60min}          & {5min}            & {43min}         \\
\bottomrule
\end{tabular}
\label{tab: partition}
\end{table*}

\section{Visualiztion}
\subsection{Additional depth Visionlization}
\label{sec: Additional depth}
\begin{figure*}[t]
    \includegraphics[width=1\linewidth]{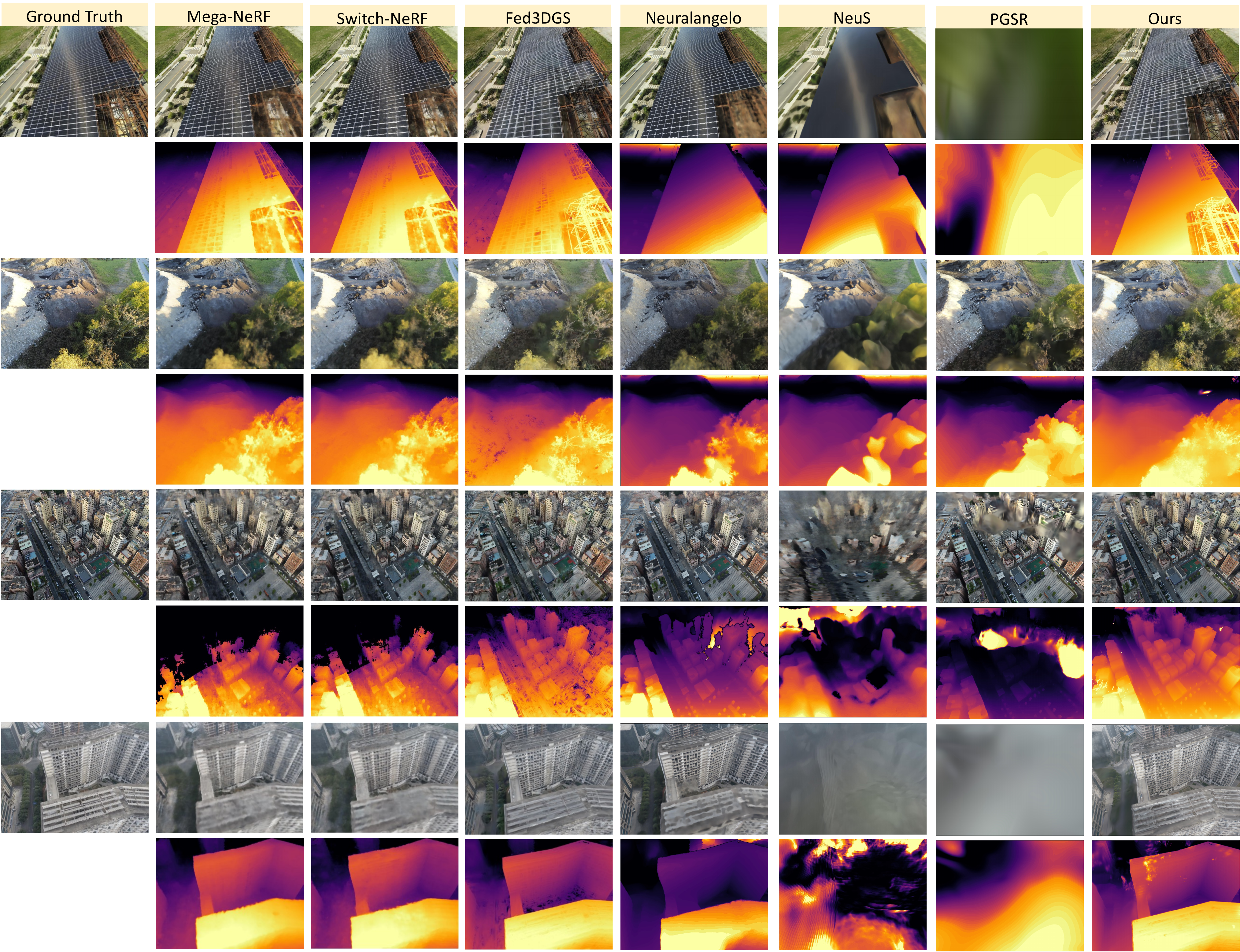}
    \caption{Qualitative results of our method and other methods in surface reconstruction datasets BlendedMVS~\cite{yao2020blendedmvs}, here we demonstrate the depth visualizations and the corresponding rendered RGB images. Images the top to bottom comes from datasets Rubble Building Residence and Campus.}
    \label{fig:depth_visual1}
\end{figure*}

The comaprison of depth maps in other ohter datasets can be seen in ~\ref{fig:depth_visual1}, It's obvious that our mehtod achieves the most accurate depth, as seen the images from Residence, we acheives the most smooth and consistant floor compare to other method.

\subsection{Additional normal Visionlization}
\label{sec: Additional normal}
\begin{figure*}[h]
    \includegraphics[width=1\linewidth]{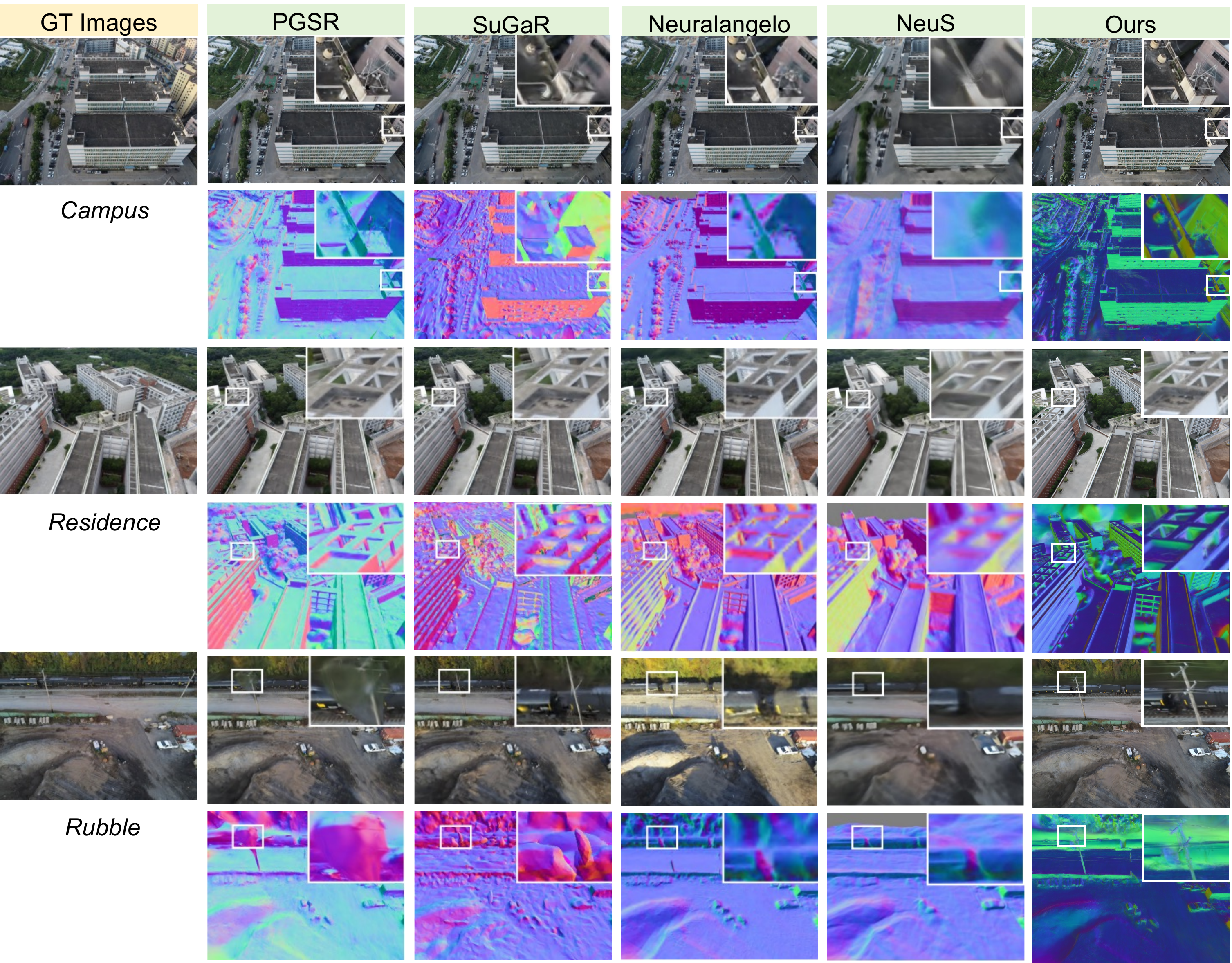}
    \caption{Qualitative results of our method and other methods in large-scale reconstruction datasets Mill-19~\cite{turki2022mega} and UrbanScene~\cite{lin2022capturing}, here we demonstrate the normal visualization and the corresponding rendered RGB images. The discriminate areas are zoomed up by '$\Box$'.}
    \label{fig:normal_visual}
\end{figure*}

In Fig ~\ref{fig:normal_visual}, We also campare our noraml maps with ohter surface reconstruction method, it's evidently that our normal map reveals more accurate details than the other method, for example the electric pole in the Rubble datasets is blur in the other methods except ours.

\begin{figure*}[t]
    \includegraphics[width=1\linewidth]{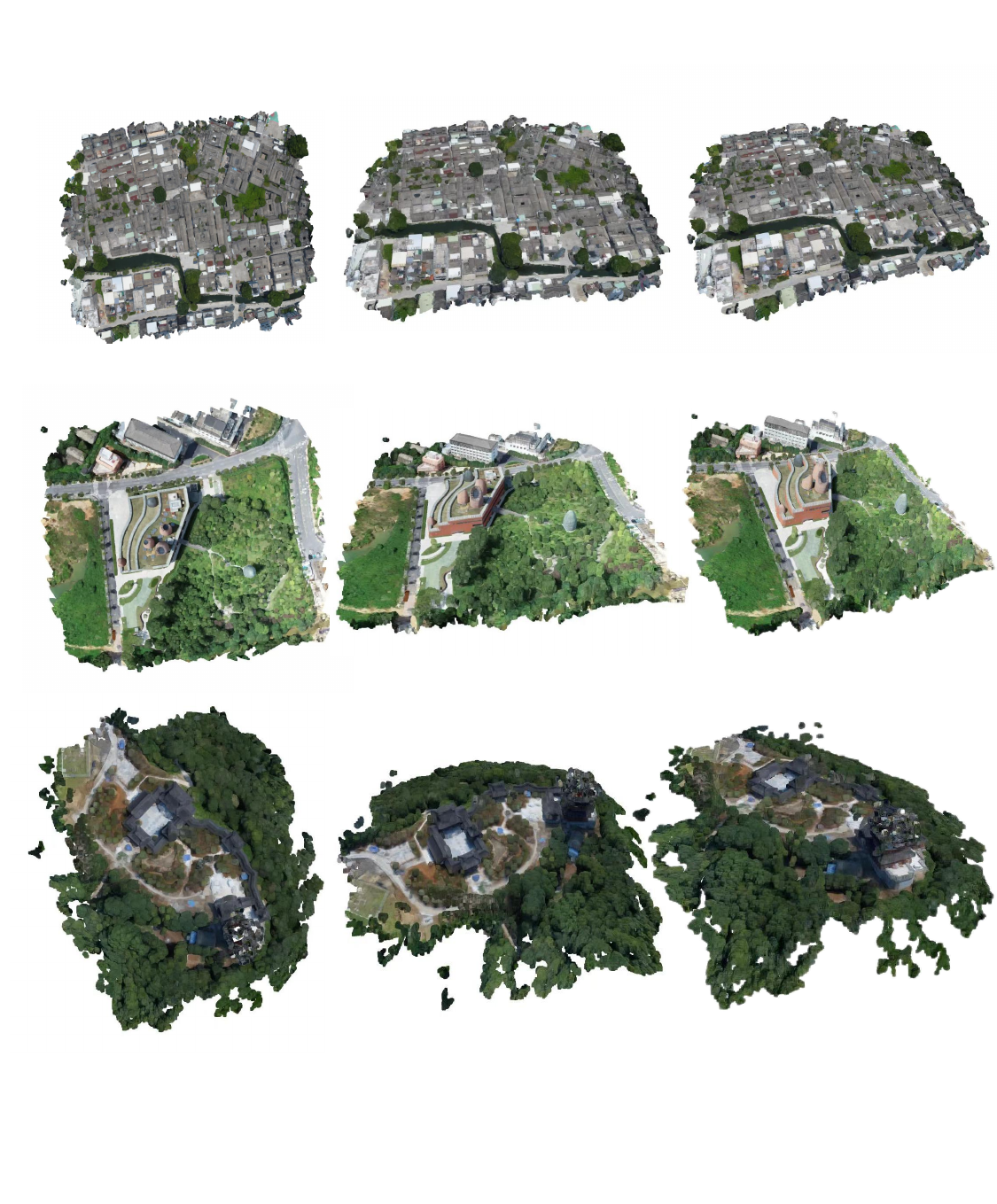}
    \caption{The full mesh map of Blendmvs Scene-01, Scene-01, Scene-03.}
    \label{figures/blend_all.pdf}
\end{figure*}


\end{document}